\documentclass[12pt]{article}
\usepackage[utf8]{inputenc}
\usepackage[T1]{fontenc}
\usepackage{graphicx,wrapfig}
\usepackage{subcaption}
\usepackage{algorithm}
\usepackage{hyperref}
\usepackage{booktabs}
\usepackage{algorithmic}
\usepackage{amsmath,amssymb,mathtools,amsthm}
\usepackage[left=1in,right=1in,top=1in,bottom=1in]{geometry}

\newcommand{\captionb}[1]{\caption{\small #1}}

\title{VQ-DRAW: A Sequential Discrete VAE}
\date{}
\author{%
  Alex Nichol \\
  \texttt{unixpickle@gmail.com}
}

\begin{document}

\maketitle

\begin{abstract}
    In this paper, I present VQ-DRAW, an algorithm for learning compact discrete representations of data. VQ-DRAW leverages a vector quantization effect to adapt the sequential generation scheme of DRAW \cite{draw} to discrete latent variables. I show that VQ-DRAW can effectively learn to compress images from a variety of common datasets, as well as generate realistic samples from these datasets with no help from an autoregressive prior.
\end{abstract}

\section{Introduction}

\begin{wrapfigure}{l}{8cm}
    \centering
    \includegraphics[width=4cm]{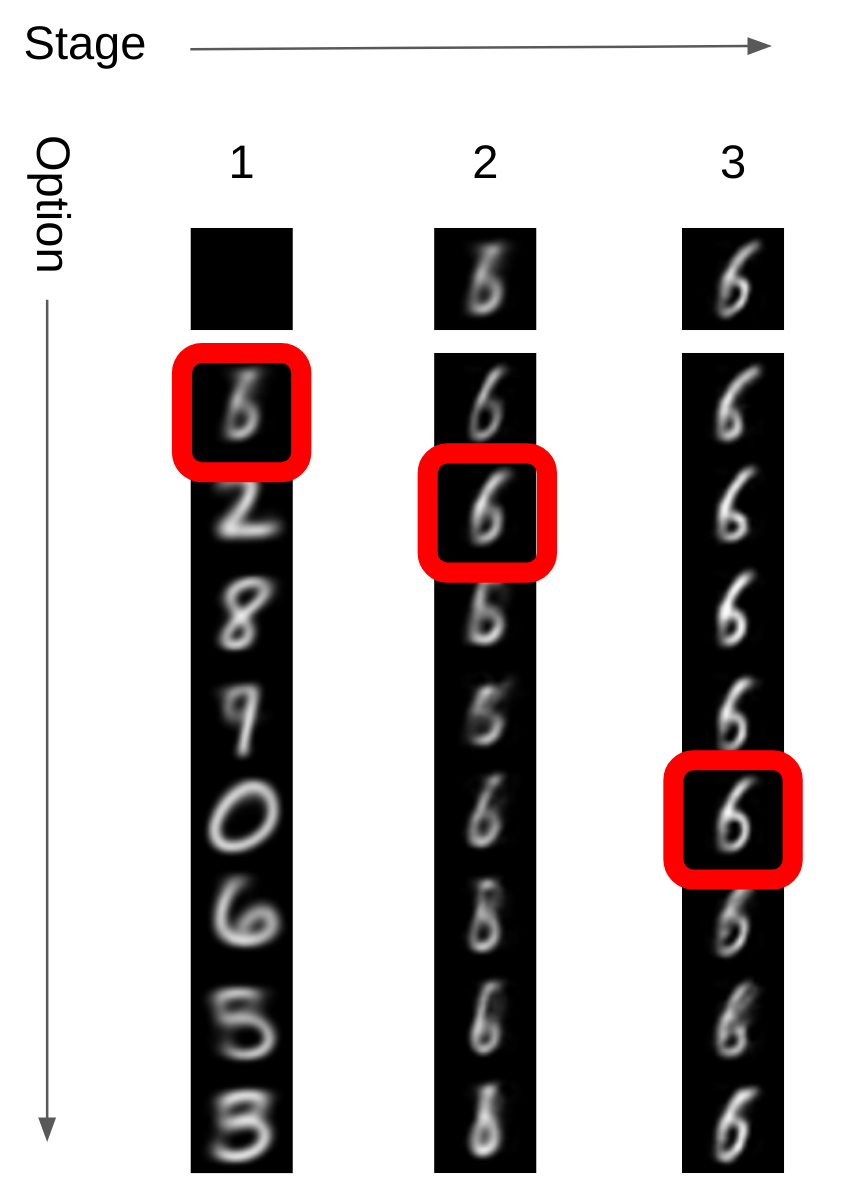}
    \captionb{\label{fig:encexample} A simplified example of VQ-DRAW encoding an MNIST digit.}
\end{wrapfigure}

This paper introduces VQ-DRAW, a new kind of discrete variational auto-encoder (VAE) \cite{vae}. Like other VAEs, VQ-DRAW aims to model a data distribution as closely as possible, enabling it to: 1) generate plausible-looking samples from the distribution, and 2) effectively compress samples from the distribution. VQ-DRAW performs well at both of these tasks, compressing images into a small fraction of their original sizes while maintaining important visual features, and generating samples which appear close to the actual data distribution. Section \ref{sec:experiments} shows examples of samples and test set reconstructions from VQ-DRAW.

Like DRAW \cite{draw}, VQ-DRAW operates in a sequential manner, generating a reconstruction in stages. Figure \ref{fig:encexample} shows how VQ-DRAW encodes an image. At each stage, a refinement network looks at the current reconstruction (the top row) and proposes variations of this reconstruction (the remaining rows). One of these variations (highlighted in red) is selected, and it is used as the reconstruction going into the next stage. The sequence of variation choices, encoded as indices, corresponds to the latent code (in this case a sequence of $\{1, 2, 5\}$). Section \ref{sec:algorithm} describes the VQ-DRAW algorithm in more detail.

VQ-DRAW can be viewed as embedding a data distribution into the leaves of a large decision tree, where each branch corresponds to another stage of encoding. After a handful of encoding stages, the encoding process has selected one leaf out of a huge number of possible leaves, whilst only traversing a tiny fraction of the entire tree.

\section{Related Work}

The variational auto-encoder (VAE) \cite{vae} is a well-known technique for deep generative modeling. One of the original insights behind VAEs was the "reparameterization trick", which makes it possible to directly optimize the variational lower bound with SGD (without using a high-variance score function estimator). However, this trick can only be applied to continuous latent variables; it cannot be used for strictly discrete latent variables.

Several works have attempted to train discrete VAEs by using continuous latent variables that gradually sharpen during training \cite{concrete} \cite{gumbel}. Other works do away with the reparameterization trick entirely, opting for other strategies for reducing variance in the variational lower bound \cite{nvil} \cite{blackbox} \cite{vimco}. However, none of these approaches bridge the performance gap between continuous and discrete VAEs.

\begin{wrapfigure}{r}{8cm}
    \centering
    \includegraphics[width=4cm]{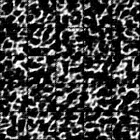}
    \captionb{\label{fig:vqvaesample} A 5x5 grid of samples from a VQ-VAE trained on MNIST, assuming a uniform prior. Latent space is 7x7x32.}
\end{wrapfigure}

More similar to VQ-DRAW, VQ-VAE \cite{vqvae} uses vector quantization to implement encoders with deterministic discrete outputs. While it is not theoretically possible to differentiate through a deterministic discrete encoder, the authors achieved good results by simply passing the gradient back through the vector quantization step. Furthermore, by assuming a uniform prior over the latents during training, they removed the need for the KL term in the variational lower bound.

In practice, however, the latent space of a VQ-VAE doesn't remain uniform. As a result, the authors needed to train strong autoregressive priors on top of the discrete latent codes to get good samples. Qualitatively, VQ-VAE models only tend to encode local "patches" of the input signal into each latent integer. As a result, random sampling of these latent codes results in outputs which are locally coherent but globally incoherent. Figure \ref{fig:vqvaesample} shows an example of this, where MNIST samples have locally coherent features (i.e. lines) but no globally coherent structure.

Like VQ-VAE, VQ-DRAW produces deterministic discrete latent codes and assumes a uniform prior. However, unlike VQ-VAE, VQ-DRAW can be locally differentiated without any gradient tricks. Furthermore, the latent codes in VQ-DRAW are not spatial, allowing VQ-DRAW to produce globally coherent samples.

Another line of related work stems from the DRAW \cite{draw} architecture. DRAW works by sequentially generating an image using attention to modify part of the image at a time. Each timestep of the sequence has its own latent code which specifies what local modification to perform. The resulting sequence of latents is treated as one larger latent code for the entire image. While DRAW itself uses attention, convolutional DRAW \cite{convdraw} maintains the same sequential architecture but opts for an attention-free approach. The sequential aspect of VQ-DRAW is similar to that of DRAW, but the latent codes in VQ-DRAW are discrete, and the neural network architectures used in VQ-DRAW are not recurrent.

\section{The VQ-DRAW Algorithm}
\label{sec:algorithm}

The VQ-DRAW algorithm encodes an input tensor as an ordered sequence of integers. Like DRAW, each element of this sequence encodes some sort of modification to the current reconstruction. Unlike DRAW, however, these latent codes are discrete, and can therefore be seen as \textit{choices} between a finite number of possible modifications to the reconstruction.

A network directly outputs all of the potential modifications to the reconstruction at each stage of encoding. The "best" modification (i.e. the one which minimizes the reconstruction error) is then chosen and passed along to the next stage. Over the course of training, the network learns to output useful variations at each stage due to a vector quantization effect, which pulls different modification outputs towards different clusters of samples in the dataset.

The variational lower bound for VQ-DRAW is simply the reconstruction error. For mean-squared error, this translates to the log-loss of an output Gaussian distribution. Like in \cite{vqvae}, the KL-divergence term normally found in the variational lower bound can be made constant by assuming a uniform prior over the latents.

\subsection{Terminology and Notation}

A single network, referred to as the \textbf{refinement network}, does the heavy lifting of both encoding and decoding. The refinement network outputs $K$ options at every stage. For stage $i$, these options are denoted as $\{R_{i,1}, ..., R_{i,K}\}$.

Both the encoding and decoding processes are sequential. Each timestep of these processes is referred to as a \textbf{stage}. For ease of notation, the variable $N$ refers to the total number of encoding/decoding stages.

The variable $X$ is used to denote a sample to be reconstructed, and $X'$ is used to denote the current reconstruction of $X$.

The loss function $\mathcal{L}$ is used to measure reconstruction error $\mathcal{L}(X', X)$.

\subsection{Encoding and Decoding}

The encoding process encodes an input tensor as a latent code of integers, $\{c_1, ..., c_N\}, c_i \in [1, K]$. The decoding process approximately reverses this transformation, taking in a latent code and producing a reconstruction tensor.

\begin{minipage}[t]{0.46\textwidth}
    \begin{algorithm}[H]
        \captionb{Encoding}
        \label{alg:encoding}
        \begin{algorithmic}
            \STATE {\bfseries Input:} $X$, tensor to be encoded.
            \STATE $X' \gets \mathbf{0}$, start with a 0 tensor.
            \FORALL{$i$ from 1 to $N$}
                \STATE ${R_{i,1}, ..., R_{i,K}} \gets \textrm{Refinement}(X')$.
                \STATE $c_i \gets \operatorname{arg\,min}_j \mathcal{L}(R_{i,j}, X)$.
                \STATE $X' \gets R_{i, c_i}$.
            \ENDFOR
            \RETURN $\{c_1, ..., c_N\}$
        \end{algorithmic}
    \end{algorithm}
\end{minipage}
\hfill
\begin{minipage}[t]{0.46\textwidth}
    \begin{algorithm}[H]
        \captionb{Decoding}
        \label{alg:decoding}
        \begin{algorithmic}
            \STATE {\bfseries Input:} $\{c_1, ..., c_N\}$, the latent code.
            \STATE $X' \gets \mathbf{0}$, start with a 0 tensor.
            \FORALL{$i$ from 1 to $N$}
                \STATE ${R_{i,1}, ..., R_{i,K}} \gets \textrm{Refinement}(X')$.
                \STATE $X' \gets R_{i, c_i}$.
            \ENDFOR
            \RETURN $X'$
        \end{algorithmic}
    \end{algorithm}
\end{minipage}

\vspace{1em}

Algorithm \ref{alg:encoding} details the encoding process. During encoding, the algorithm keeps track of the current reconstruction, which always starts out the same (usually as some tensor of 0s). At each stage of encoding, the refinement network proposes $K$ variations of the current reconstruction. The variation with the lowest reconstruction error is chosen, and the index of this variation is saved as a latent component. After $N$ stages of encoding, we have all the latent components $\{c_1, ..., c_N\}, c_i \in [1, K]$, forming the complete latent code. Thus, the latent code is $N \cdot log_2(K)$ bits.

Decoding proceeds in a similar fashion, as shown in Algorithm \ref{alg:decoding}. For each decoding step $i$ from 1 to $N$, the current reconstruction is fed into the refinement network, and the variation at index $c_i$ becomes the new current reconstruction. After all $N$ stages are performed, the current reconstruction is returned.

It should be noted that the encoding process keeps track of the current reconstruction, and therefore yields the reconstruction for free (i.e. with no extra compute). Thus, during training, no extra decoding step has to be performed.

\subsection{Training}

It is possible to backpropagate through the encoding process under the assumption that $\{c_1, ..., c_N\}$ remain fixed. In other words, the final reconstruction error is locally smooth and differentiable, and we can use SGD to minimize it. However, the gradient for a given sample will ignore most of the outputs of the refinement network, since a specific refinement is selected at each stage of the encoding process and the remaining refinements are discarded.

In particular, at stage $i$, the refinement network outputs refinements $\{R_{i,1}, ..., R_{i,k}\}$, but only one refinement $R_{i,c_i}$ is carried on to the next stage of encoding. The other refinements are not used and do not directly contribute to the final reconstruction loss. Furthermore, if refinement $R_{i,j}$ is never used for any sample in the dataset (perhaps because it has a bad set of initial biases), then there will be no gradient signal to directly improve $R_{i,j}$ so that it can be used for encodings later on in training.

A slightly modified loss function can alleviate the above concern. This loss function aims to minimize not only the reconstruction errors for the \textit{chosen} reconstructions (Equation \ref{eq:lossrecon}), but also for \textit{all} of the reconstructions, even the unchosen ones (Equation \ref{eq:lossall}). Thus, if a refinement option $R_{i,j}$ is never being used, it will gradually be pulled closer to the real refinement distribution until it finally is used to encode some sample in the dataset. The final loss is shown in Equation \ref{eq:loss}.

\begin{alignat}{2}
    \ell_{chosen} &= \frac{1}{N} \sum_{i=1}^N \mathcal{L}(R_{i, c_i}, X) \label{eq:lossrecon} \\
    \ell_{all} &= \frac{1}{NK} \sum_{i=1}^N \sum_{j=1}^K \mathcal{L}(R_{i,j}, X) \label{eq:lossall} \\
    \ell_{total} &= \ell_{chosen} + \alpha \cdot \ell_{all}\textrm{, }\alpha \ll 1 \label{eq:loss}
\end{alignat}

I experimented with several modifications to Equation \ref{eq:loss} that did not pan out. One natural idea would be to remove the $\ell_{all}$ loss term entirely, and instead add some exploration by choosing $c_i$ randomly with some small probability $\epsilon$. I found that this reduced both training and test performance, even with randomness turned off during evaluation. I also tried replacing $\ell_{chosen}$ with $\mathcal{L}(R_{N, c_N}, X)$, minimizing final reconstruction error while ignoring intermediate reconstruction errors. This modification prevented the model from converging, even though it is technically closer to the actual objective of the encoding process.

\subsection{Network Architecture}

The refinement network must take in a reconstruction and produce $K$ refinements to this reconstruction. There are plenty of architectures that could serve this purpose, but I found three things to be universally beneficial regardless of architecture:

\begin{enumerate}
    \item \textbf{Stage conditioning:} training converges to much better solutions when the refinement network receives the stage index $i$ as an extra input. Presumably, this allows the network to perform different sorts of refinements at different stages, focusing on more fine-grained details at later stages of encoding. Stage conditioning resulted in large improvements on CIFAR-10, where the refinement sequence is an order of magnitude longer than the sequences for MNIST and SVHN.
    \item \textbf{Segmented models:} this is a special kind of stage conditioning where entirely different refinement network parameters are used for different segments of stages. For example, the SVHN experiments use a different network for each segment of 5 stages, so stages 1-5 are refined by a totally different set of network parameters than stages 5-10. This drastically improves reconstruction losses, which may either be caused by 1) the sheer number of extra parameters, or 2) the refinement network being able to focus on different aspects of the images at different parts of the encoding process. The latter seems plausible, given the success of stage conditioning in general.
    \item \textbf{Residual outputs:} refinements are output as residuals to be added to the current reconstruction. In other words, the network itself outputs deltas $\{\Delta_1, ..., \Delta_k\}$ which are added to the current reconstruction $X'$ to produce refinements $\{X' + \Delta_1, ..., X' + \Delta_K\}$. This is motivated by the fact that roughly half of the deltas should have a positive dot product with the gradient of the reconstruction loss, so half of the refinements should be improvements (given small enough deltas).
\end{enumerate}

For images, I use a CNN refinement network architecture which downsamples the image, processes it with numerous residual layers, and then performs transposed convolutions to upsample to the original resolution. A final convolutional head produces $K \times C$ channels, where $C$ is the number of channels in the reconstructed images. Thus, most of the architecture does not grow with $K$, only the final layer.

To condition CNN models on the stage index, I introduce layers after every convolution that multiply the channels by a per-stage mask vector. In particular, for every stage $i$ and convolutional layer $l$ with $C$ output channels, there is a different mask $M_{l,i}$ of length $C$ which is broadcasted and multiplied by the outputs of the convolutional layer.

In most of my CNN models, I use group normalization \cite{groupnorm} after every ReLU non-linearity. I chose group normalization instead of batch normalization \cite{batchnorm} mainly because group normalization makes it simpler to use bigger batches via gradient accumulation.

\section{Results}
\label{sec:experiments}

\begin{figure}[t]
    \centering
    \begin{subfigure}{0.4\textwidth}
        \centering
        \includegraphics[width=\textwidth]{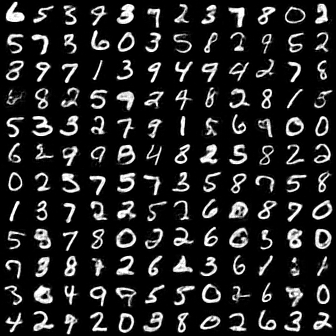}
        \captionb{MNIST}
    \end{subfigure}%
    \hspace{1em}%
    \begin{subfigure}{0.4\textwidth}
        \centering
        \includegraphics[width=\textwidth]{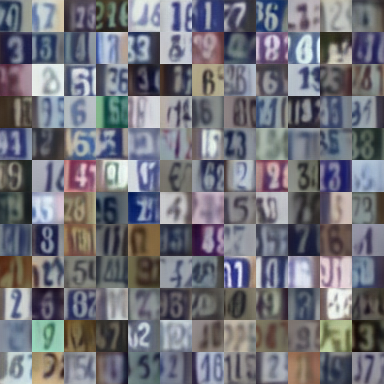}
        \captionb{SVHN}
    \end{subfigure}
    \par\bigskip
    \begin{subfigure}{0.4\textwidth}
        \centering
        \includegraphics[width=\textwidth]{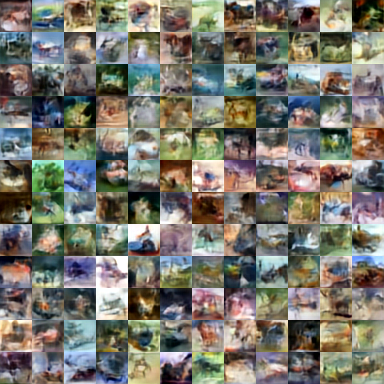}
        \captionb{CIFAR-10}
    \end{subfigure}%
    \hspace{1em}%
    \begin{subfigure}{0.4\textwidth}
        \centering
        \includegraphics[width=\textwidth]{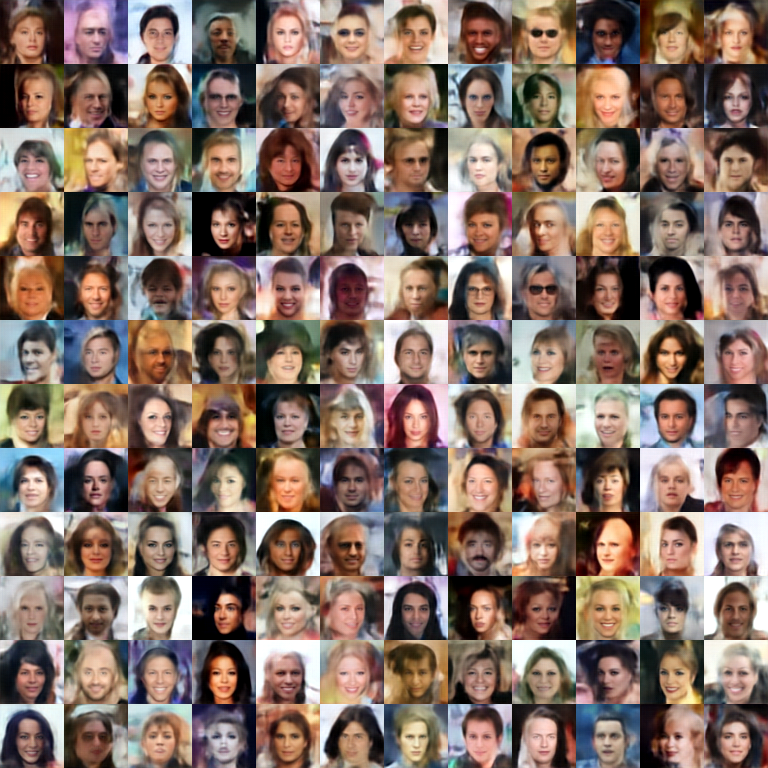}
        \captionb{CelebA}
    \end{subfigure}
    \hfill
    \captionb{\label{fig:samples} Generated samples from trained VQ-DRAW models.}
\end{figure}

I trained VQ-DRAW models on four different image datasets: MNIST \cite{mnist}, SVHN \cite{svhn}, CIFAR-10 \cite{cifar10}, and CelebA \cite{CelebA}. The former two datasets are ideal for rapid iteration and experimentation, while the latter two are useful for confirming that ideas scale to harder datasets. \footnote{These experiments were all run on single machines with one GPU. This means that anybody with access to consumer-grade deep learning hardware should be capable of reproducing, modifying, and ablating these experiments in a reasonable amount of time.}

Table \ref{tbl:latentsize} specifies the hyperparameters used for training. All models were optimized with Adam \cite{adam} using a fixed step size of $10^{-3}$. To train the CIFAR-10 and CelebA models on a single GPU, I used gradient accumulation and gradient checkpointing \cite{gradcheckpoint1} \cite{gradcheckpoint2}.

\begin{table}[h]
    \centering
    \begin{tabular}{rllll}
        \toprule
        & MNIST & SVHN & CIFAR-10 & CelebA \\
        \midrule
        Options ($K$) & 64 & 64 & 64 & 64 \\
        Stages ($N$) & 10 & 20 & 100 & 100 \\
        Stages per segment & 10 & 5 & 10 & 10 \\
        Unchosen weight ($\alpha$) & 0.01 & 0.01 & 0.01 & 0.01 \\
        Batch size & 32 & 512 & 512 & 512 \\
        Steps & 50,000 & 4,375 & 2,175 & 2,262 \\
        \midrule
        Image size & 28x28x1 & 32x32x3 & 32x32x3 & 64x64x3 \\
        Code size (bits) & 60 & 120 & 600 & 600 \\
        \bottomrule
    \end{tabular}
    \captionb{\label{tbl:latentsize} Experimental hyperparameters..}
\end{table}

\begin{figure}[h]
    \centering
    \begin{subfigure}{0.45\textwidth}
        \centering
        \includegraphics[width=\textwidth]{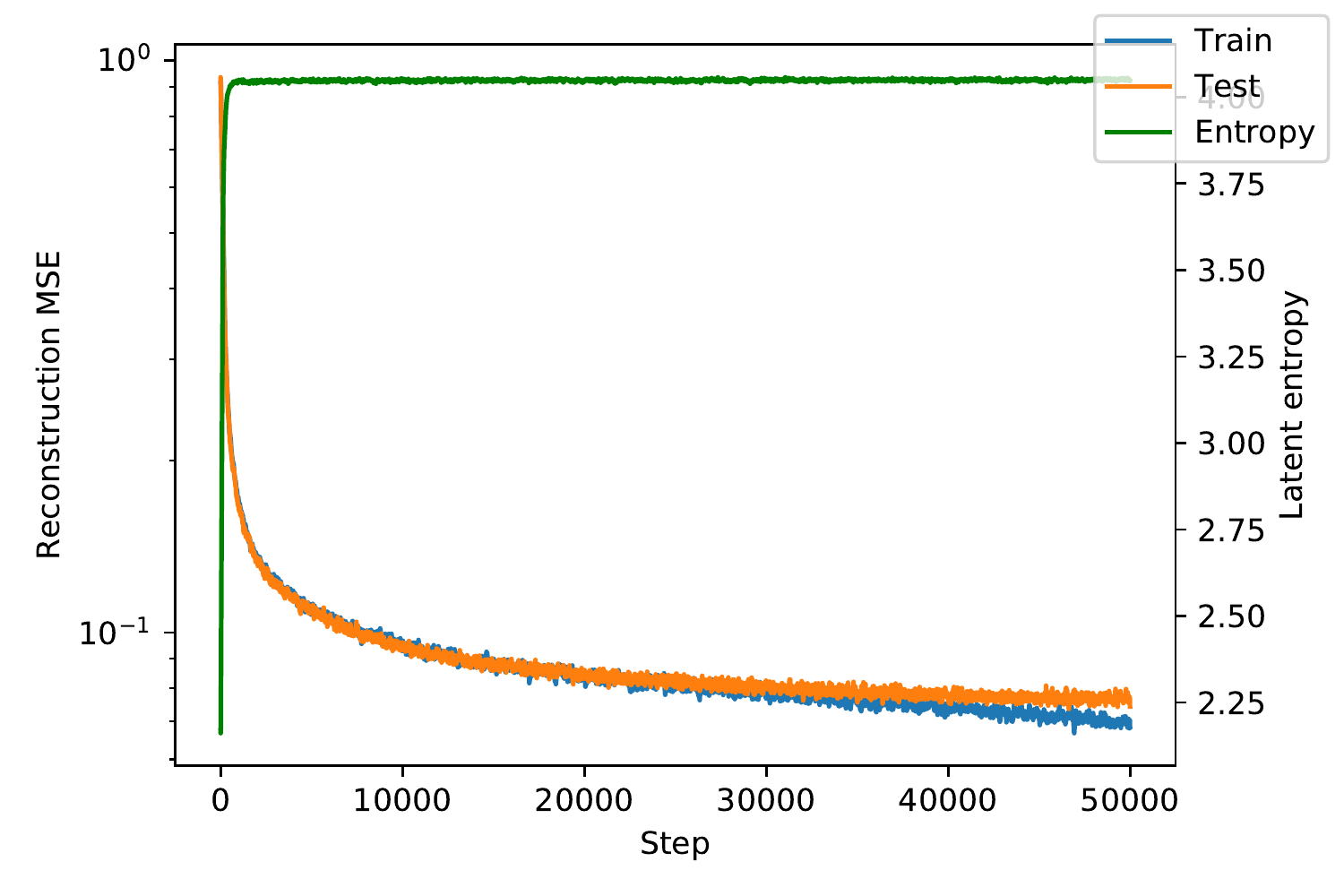}
        \captionb{MNIST}
    \end{subfigure}
    \begin{subfigure}{0.45\textwidth}
        \centering
        \includegraphics[width=\textwidth]{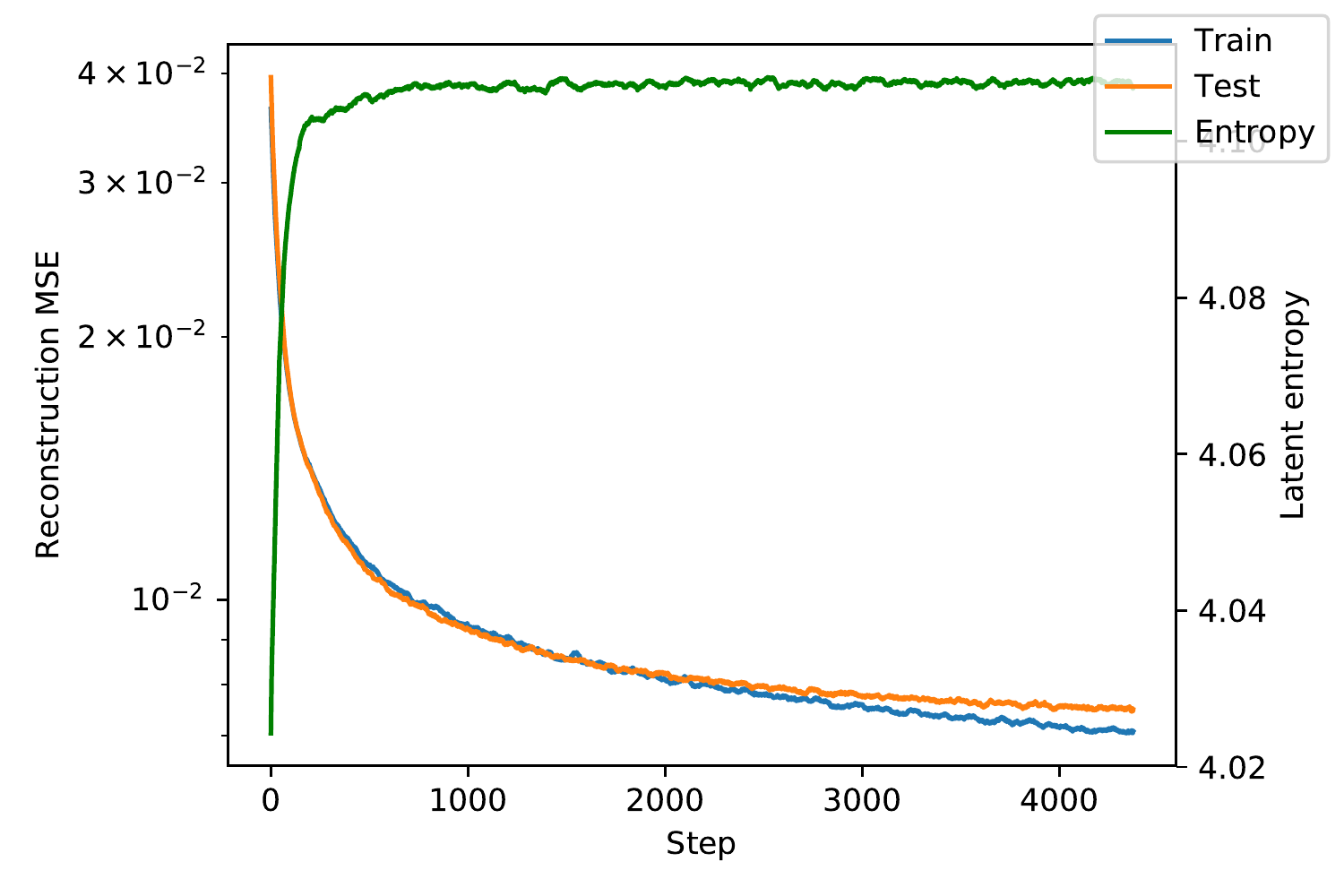}
        \captionb{SVHN}
    \end{subfigure}
    \begin{subfigure}{0.45\textwidth}
        \centering
        \includegraphics[width=\textwidth]{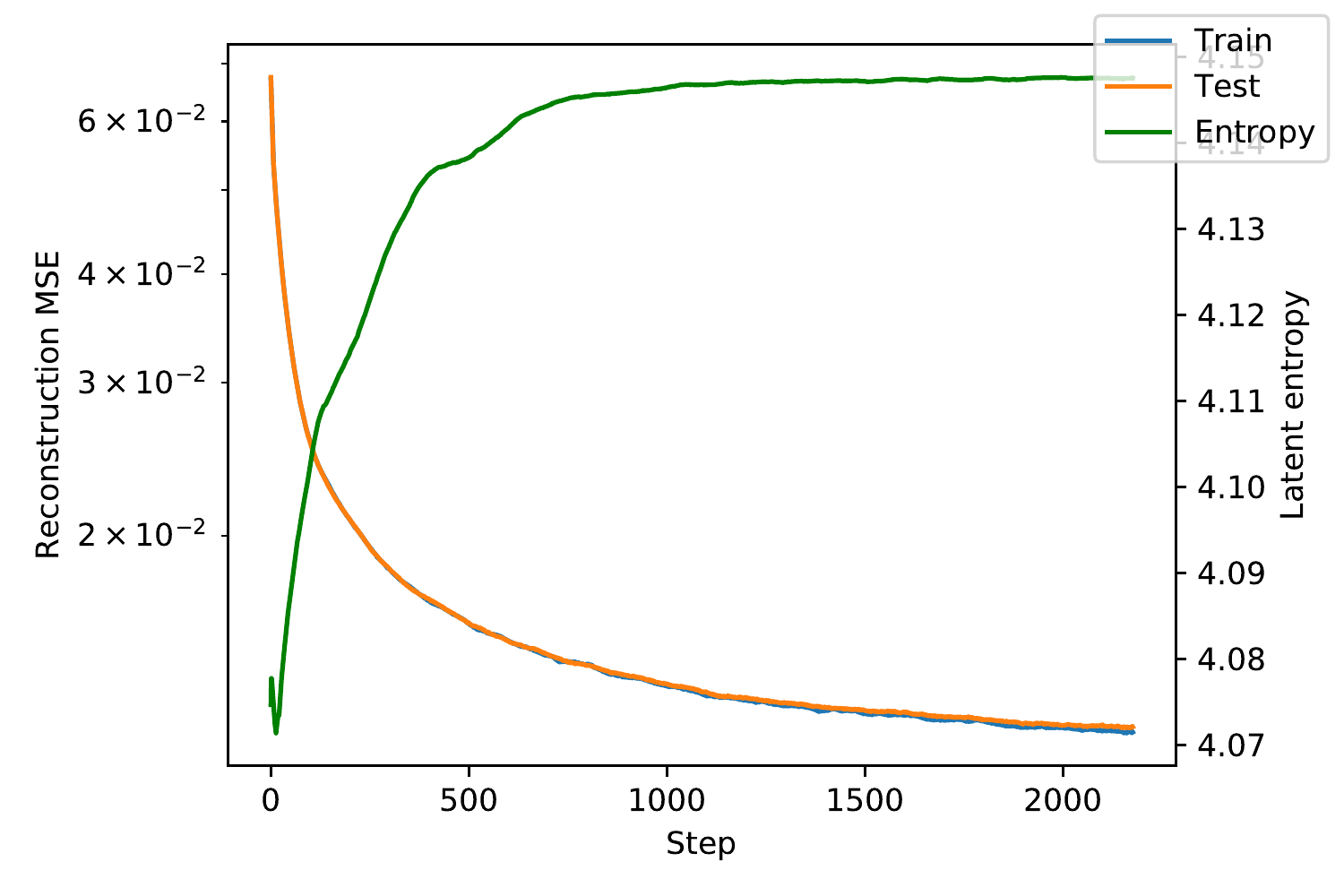}
        \captionb{CIFAR-10}
    \end{subfigure}
    \begin{subfigure}{0.45\textwidth}
        \centering
        \includegraphics[width=\textwidth]{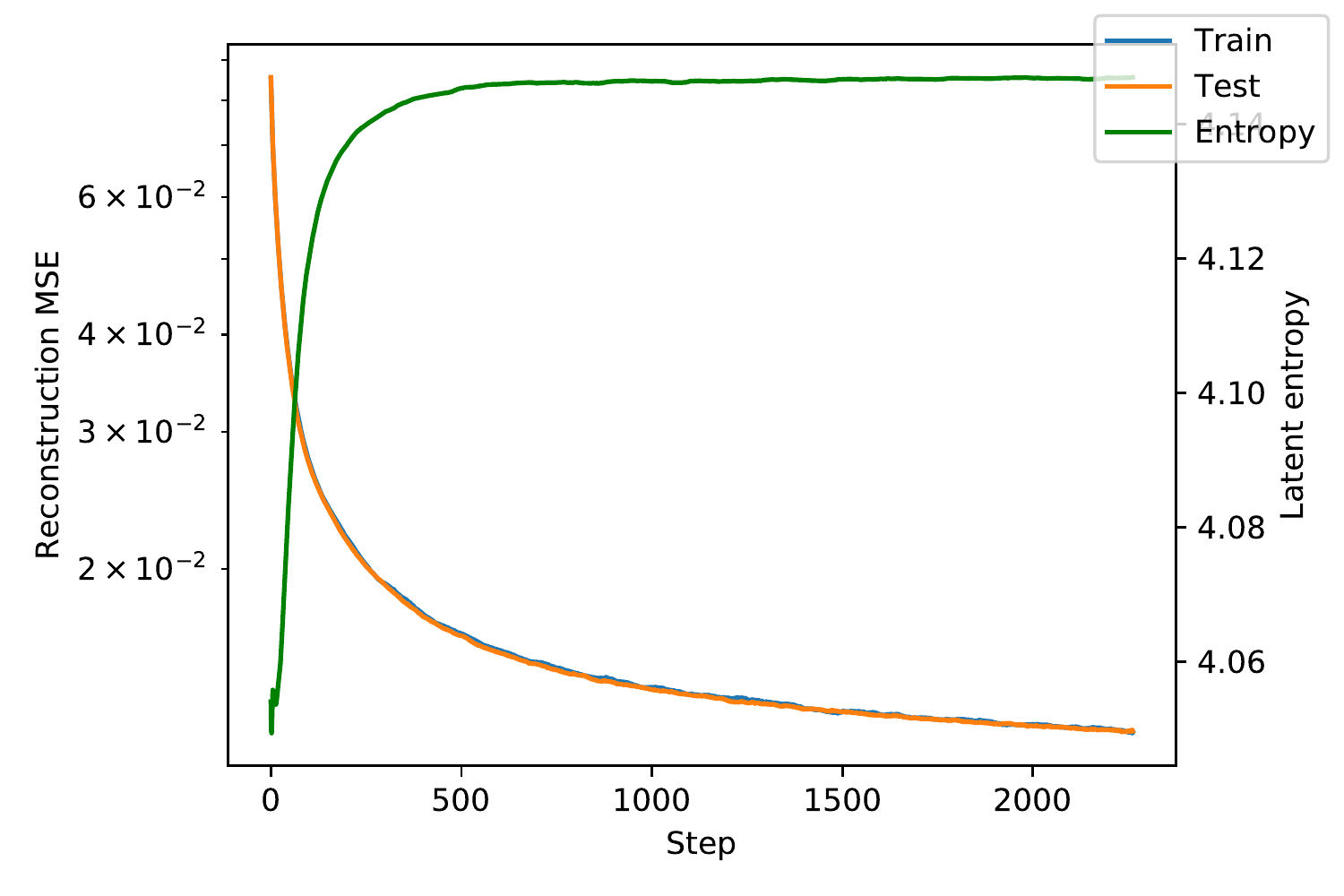}
        \captionb{CelebA}
    \end{subfigure}
    \hfill
    \captionb{\label{fig:curves} Learning curves on four different image datasets.}
\end{figure}

Figure \ref{fig:curves} shows the learning curves for VQ-DRAW models trained on all four datasets. The plots also include the entropy of the latent codes (per mini-batch) to show how well the model spreads out its usage of different refinement options. This metric is not perfect, but it does reveal when a model is not using many of its refinement options effectively. As is apparent in all of these experiments, entropy quickly rises and then remains near the maximum value of $ln(64) \approx 4.16$. However, it cannot reach this value when the mini-batch size is finite.

\begin{figure}[h]
    \centering
    \begin{subfigure}{0.3\textwidth}
        \centering
        \includegraphics[width=\textwidth]{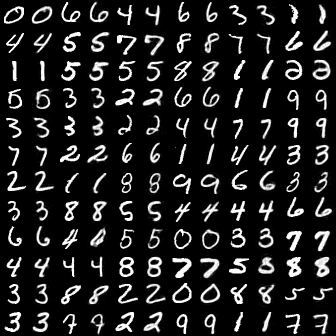}
        \captionb{MNIST}
    \end{subfigure}%
    \hspace{1em}%
    \begin{subfigure}{0.3\textwidth}
        \centering
        \includegraphics[width=\textwidth]{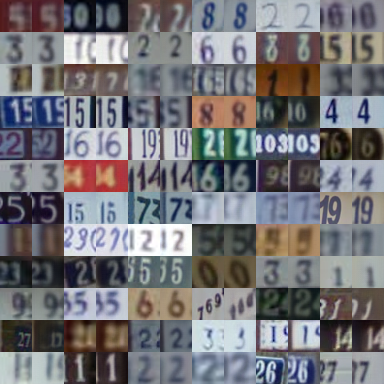}
        \captionb{SVHN}
    \end{subfigure}
    \par\bigskip
    \begin{subfigure}{0.3\textwidth}
        \centering
        \includegraphics[width=\textwidth]{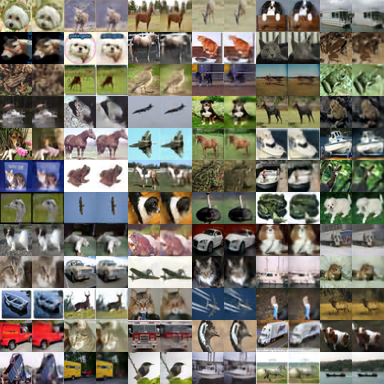}
        \captionb{CIFAR-10}
    \end{subfigure}%
    \hspace{1em}%
    \begin{subfigure}{0.3\textwidth}
        \centering
        \includegraphics[width=\textwidth]{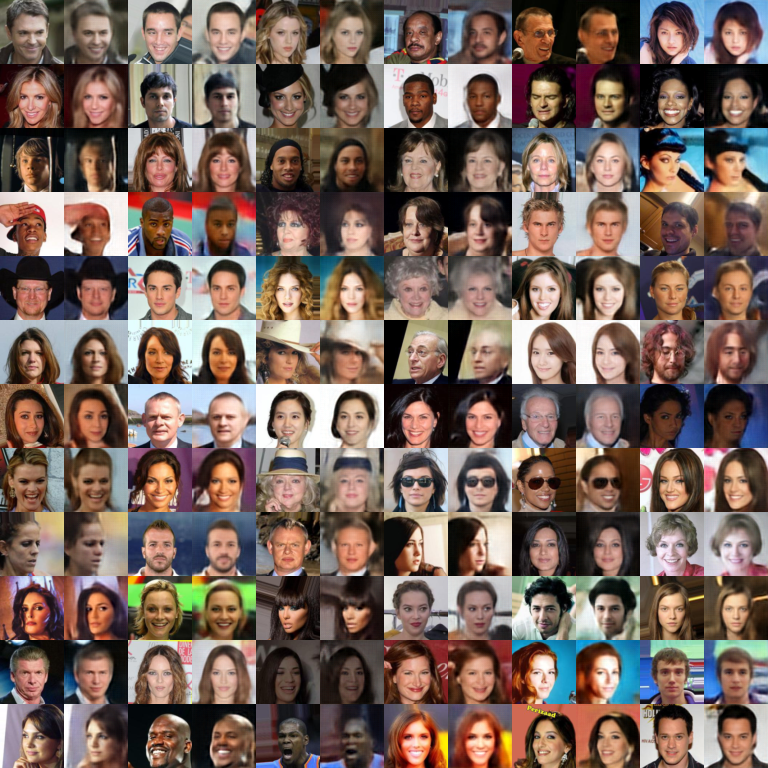}
        \captionb{CelebA}
    \end{subfigure}
    \hfill
    \captionb{\label{fig:recon} Test set reconstructions from trained VQ-DRAW models on four different datasets.}
\end{figure}

To qualitatively evaluate how well the models have actually learned their data distributions, it is helpful to look at samples and reconstructions after training. Figure \ref{fig:samples} shows random samples produced by these models, and Figure \ref{fig:recon} shows reconstructions from randomly chosen test set images.

I did not attempt to approximate log-likelihoods for VQ-DRAW. It is likely that the variational lower bound would be very loose for VQ-DRAW models, and this looseness would depend on the hyperparameter $N$ (the number of stages). To see why, imagine increasing $N$ to infinity. At the limit, each additional stage will increase the variational lower bound by a constant amount while leaving the log-likelihood unchanged, since each additional stage will have no reason to modify the already-perfect reconstructions (and will thus add no new information). To get better estimates of log-likelihood, a tighter bound (perhaps with some degree of marginalization over latent codes) should be used. Of course, it would also be necessary to modify the way VQ-DRAW outputs predictions at each stage: either a standard deviation should be produced as well as a mean (for continuous outputs), or the output pixels should be discretized and predicted with a softmax.

\section{Future Work}

While this paper only applies VQ-DRAW to images, it is theoretically possible to apply VQ-DRAW to text, audio, 3D models, and many other types of data. Experimenting with these data modalities could shed light on how VQ-DRAW works and where its strengths and weaknesses lie.

It should of course be noted that the experiments in this paper were limited by computational constraints. With larger models, bigger batch sizes, and more training iterations, the same core algorithm may achieve significantly better results.

Aside from more compute, there are a number of potential modifications to VQ-DRAW that might improve sample quality, reconstruction error, and/or computational efficiency. Here are a few examples:

\begin{itemize}
    \item The VQ-DRAW algorithm, as presented here, uses a greedy choice at each stage of encoding. Perhaps VQ-DRAW could be improved by using beam search or some other more intelligent criterion for choosing proposals from the refinement network.
    \item The refinement network could be made "progressive" \cite{progressivegan} \cite{vqvae2}, producing lower resolution outputs at earlier stages and higher resolution outputs at later stages.
    \item Unlike feed-forward VAE models, the computational cost of VQ-DRAW depends on the number of encoding stages. Thus, training VQ-DRAW on very complex data with tens of thousands of latent bits may be computationally difficult. To alleviate this dilemma, it may be beneficial to periodically distill segments of the encoding process into a single feed-forward network, essentially using the latent encodings from VQ-DRAW as supervised targets for a more efficient model.
\end{itemize}

It remains to be seen how the learned representations from VQ-DRAW could be used for downstream tasks. I briefly experimented with classifying MNIST digits from VQ-DRAW latent codes, and found discouraging results. Perhaps a modified VQ-DRAW objective could improve these results.

Recent work in unsupervised learning, such as CPC \cite{cpc} and SimCLR \cite{simclr}, do not rely on reconstructions at all. Instead, these methods aim to directly encode the mutual information between different patches (or augmentations) of the same image. This approach to unsupervised learning has been very fruitful for images, where reconstructions can focus too much on unimportant features. It may be possible to adapt the principles of VQ-DRAW to a contrastive predictive framework, producing useful discrete latent codes for downstream classification tasks.

\newpage

\bibliographystyle{IEEEtran}
\bibliography{IEEEabrv,main}

\newpage

\appendix

\section{Stages of Encoding}

\begin{figure}[h]
    \centering
    \begin{subfigure}{0.3\textwidth}
        \centering
        \includegraphics[width=\textwidth]{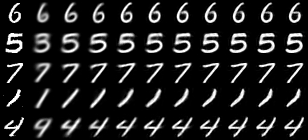}
        \captionb{MNIST}
    \end{subfigure}%
    \hspace{1em}%
    \begin{subfigure}{0.55\textwidth}
        \centering
        \includegraphics[width=\textwidth]{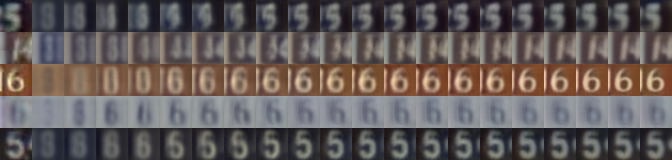}
        \captionb{SVHN}
    \end{subfigure}
    \par\bigskip
    \begin{subfigure}{0.55\textwidth}
        \centering
        \includegraphics[width=\textwidth]{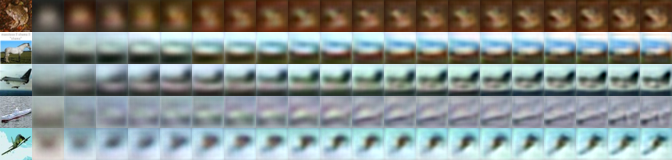}
        \captionb{CIFAR-10}
    \end{subfigure}%
    \hspace{1em}%
    % \begin{subfigure}{0.55\textwidth}
    %     \centering
    %     \includegraphics[width=\textwidth]{celeb_stages_small.png}
    %     \captionb{CelebA}
    % \end{subfigure}
    \hfill
    \captionb{\label{fig:stages} The first 20 (or fewer) stages during encoding for random test images. Left-most column is the target image. CelebA samples are not included due to file-size constraints.}
\end{figure}

\end{document}